\documentclass[letterpaper]{article} 
\usepackage{aaai2026}  
\usepackage{times}  
\usepackage{helvet}  
\usepackage{courier}  
\usepackage[hyphens]{url}  
\usepackage{graphicx} 
\usepackage{natbib}  
\usepackage{caption} 
\usepackage{algorithm}
\usepackage{algorithmic}
\usepackage{amsmath, amssymb}
\usepackage{multirow}
\usepackage{booktabs}
\usepackage{subfigure}
\usepackage{tikz}
\usepackage{xfrac}
\usepackage[table]{xcolor}

\usepackage{xspace}
\newcommand{\ie}{{\emph{i.e.}}\xspace}
\newcommand{\eg}{{\emph{e.g.}}\xspace}

\newcommand{\downrightarrow}{\hspace{3pt}\raisebox{1.5pt}{\begin{tikzpicture}[scale=0.4, baseline=(current bounding box.south)]
    \draw[-latex] (0.0, 0.5) -- (0.0, 0.25) -- (0.7, 0.25);
  \end{tikzpicture}} }

\usepackage{newfloat}
\usepackage{listings}
\DeclareCaptionStyle{ruled}{labelfont=normalfont,labelsep=colon,strut=off} 
\floatstyle{ruled}
\newfloat{listing}{tb}{lst}{}
\floatname{listing}{Listing}
\title{Real-Time 3D Object Detection with Inference-Aligned Learning}
\author{
Chenyu Zhao\textsuperscript{\rm 1}, Xianwei Zheng\textsuperscript{\rm 1}\thanks{Corresponding author.}, Zimin Xia\textsuperscript{\rm 2},  Linwei Yue\textsuperscript{\rm 3}, Nan Xue\textsuperscript{\rm 4}\\
}
\affiliations{
    \textsuperscript{\rm 1}The State Key Lab. LIESMARS, Wuhan University \\
    \textsuperscript{\rm 2}École polytechnique fédérale de Lausanne (EPFL) \\ 
    \textsuperscript{\rm 3}School of Geography and Information Engineering, China University of Geosciences\\ 
    \textsuperscript{\rm 4}Ant Group\\
    \{cyzhao, zhengxw\}@whu.edu.cn, zimin.xia@epfl.ch, yuelw@cug.edu.cn, xuenan@ieee.org
}

\usepackage{bibentry}

\begin{document}

\maketitle

\begin{abstract}
Real-time 3D object detection from point clouds is essential for dynamic scene understanding in applications such as augmented reality, robotics, and navigation. 
We introduce a novel Spatial-prioritized and Rank-aware 3D object detection (SR3D) framework for indoor point clouds, to bridge the gap between how detectors are trained and how they are evaluated. 
This gap stems from the lack of spatial reliability and ranking awareness during training, which conflicts with the ranking-based prediction selection used at inference.
Such a training-inference gap hampers the model’s ability to learn representations aligned with inference-time behavior.
To address the limitation, SR3D consists of two components tailored to the spatial nature of point clouds during training: a novel spatial-prioritized optimal transport assignment that dynamically emphasizes well-located and spatially reliable samples, and a rank-aware adaptive self-distillation scheme that adaptively injects ranking perception via a self-distillation paradigm.
Extensive experiments on ScanNet V2 and SUN RGB-D show that SR3D effectively bridges the training-inference gap and significantly outperforms prior methods in accuracy while maintaining real-time speed.
Code is publicly available at https://github.com/zhaocy-ai/sr3d. 
\end{abstract}

\section{Introduction}
With the increasing availability of 3D sensing technologies, understanding 3D point clouds has become a crucial task in computer vision. 
We are interested in {\em 3D object detection for point clouds of indoor scenes}, aiming to localize 3D bounding boxes and determine their semantic classes in real time. 
Robust and real-time 3D object detection is vital for holistic and dynamic scene understanding, enabling critical applications in augmented reality, embodied robotics, and navigation.
In such scenarios, the perception system must process input point clouds and make decisions within tight time constraints to ensure responsiveness and safety.

\begin{figure}[t]
    \centering
    \includegraphics[width=0.9\linewidth]{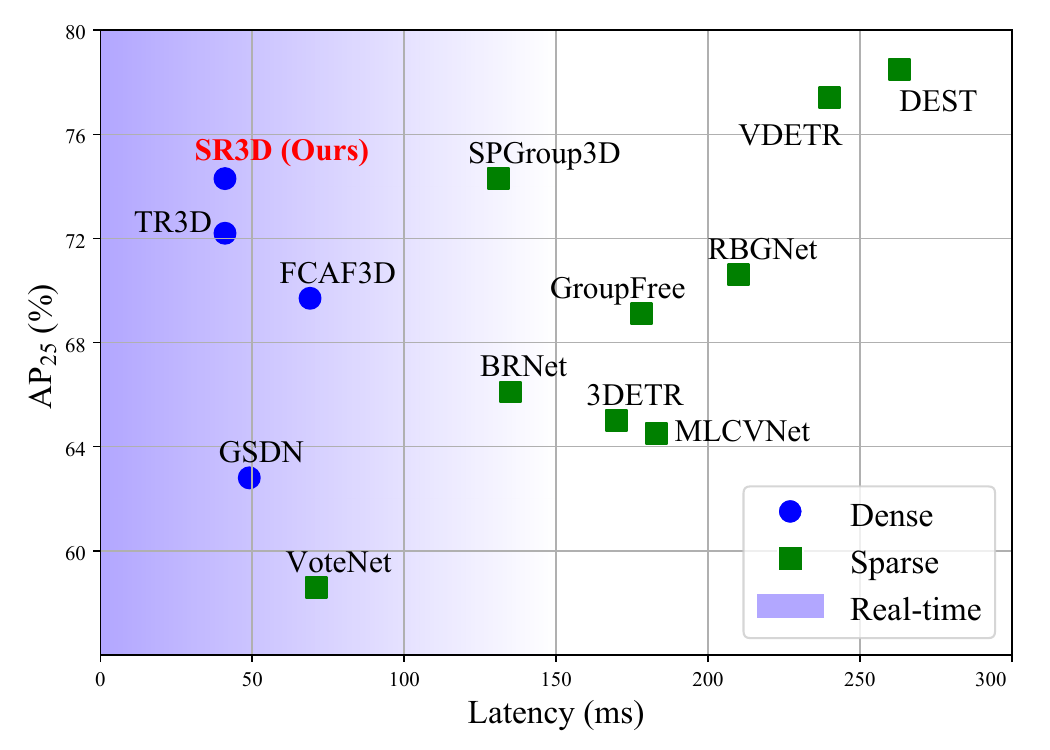}
    \caption{AP$_{25}$ vs. latency on the ScanNet V2 validation set. 
    Our proposed SR3D achieves accurate and fast detection from indoor point clouds. 
    Latency is measured on a single RTX 4090 GPU. 
    The metrics AP$_{25}$ is mean Average Precision under the IoU threshold of 0.25.}
    \label{real-time}
\end{figure}

Current 3D detectors are broadly divided into sparse and dense detection paradigms based on their proposal generation mechanisms.
Sparse detectors~\cite{qi2019deep, misra2021end} rely on refining sparse proposals, which incur high memory costs and limit scalability, making them less suitable for real-time applications.
As shown in Fig.~\ref{real-time}, dense detectors~\cite{gwak2020generative, rukhovich2022fcaf3d} are more efficient, as they utilize anchors densely to cover objects and predict bounding boxes with semantic labels in a single pass.
Thus, we adopt dense detection frameworks as they are better aligned with the real-time requirements. 

\begin{figure*}[t]
    \centering
    \subfigure[Spatial reliability absence]{
    \includegraphics[width=0.4\linewidth]{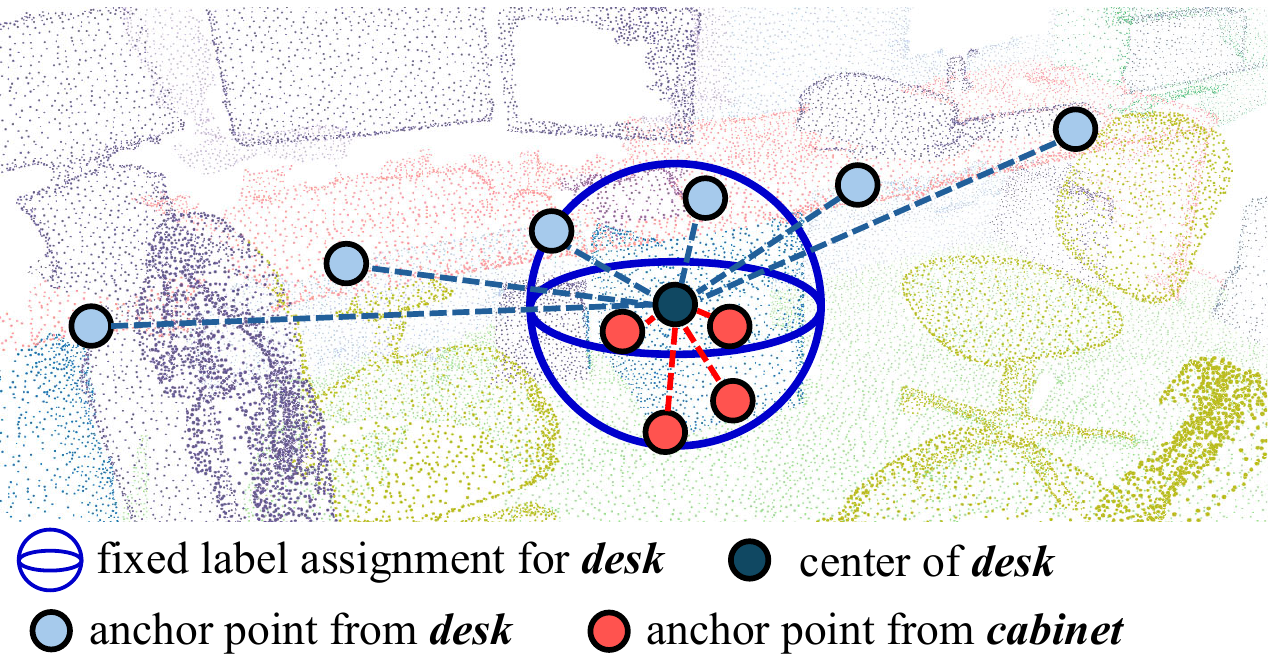}}
    \subfigure[Ranking awareness absence]{
    \includegraphics[width=0.5\linewidth]{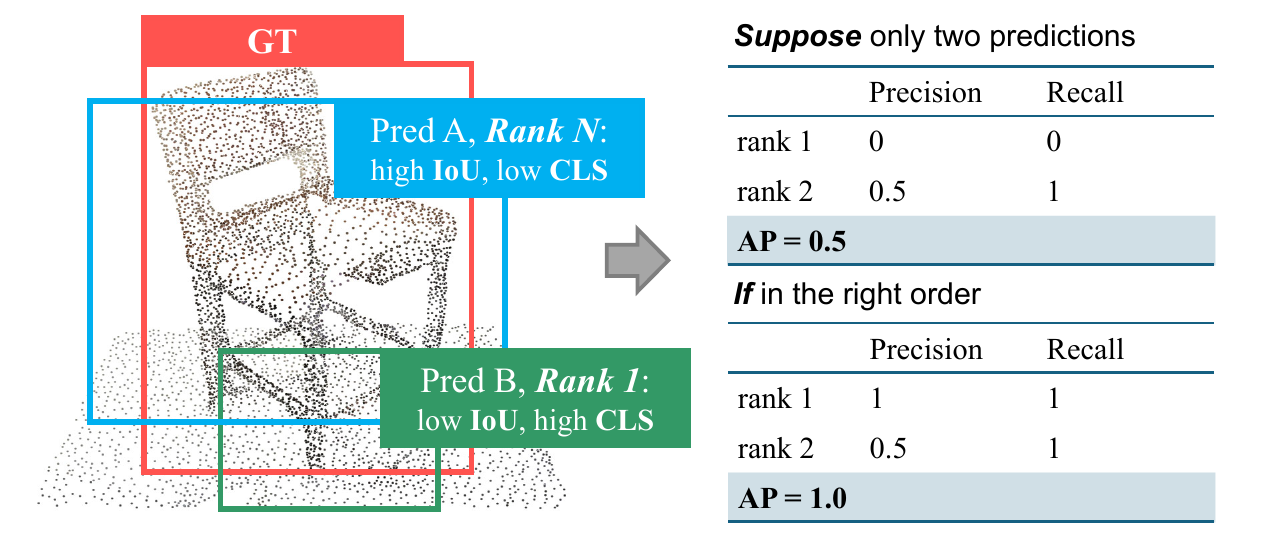}}
    \caption{Illustration of core limitations in current dense 3D detectors. (a) Fixed heuristic label assignment misidentifies high-quality anchors for the \textit{desk}, being misled by spatial clutter in the indoor scene. (b) Rank-agnostic supervision leads to incorrect ranking of \textit{chair} predictions, degrading performance under Average Precision (AP) evaluation. We use 2D boxes for simplicity.}
    \label{limitation}
\end{figure*}

However, a fundamental problem in current dense 3D object detectors lies in their inability to align training supervision with inference behavior: \textit{predictions are selected and optimized without considering either their spatial reliability or relative ranking} during training, as shown in Fig.~\ref{limitation}.
This misalignment ultimately hinders the detector's ability to learn robust and discriminative representations.

To address the issue, we propose a simple but effective framework SR3D tailored for real-time 3D object detection that explicitly integrates the awareness of spatial reliability and ranking into the supervision process.
First, we observe that detectors often fail to focus on the most informative samples, especially when dealing with occluded or geometrically ambiguous objects.
This is largely due to the use of fixed heuristic label assignment strategies, such as center priors and IoU thresholds, which ignore the actual spatial reliability of anchors during training.
These strategies thereby overlook important and various geometric cues inherent in 3D objects and often mislead the detector to suboptimal solutions.
Thus, we introduce the Spatial-Prioritized Optimal Transport Assignment (SPOTA), a formulation that casts label assignment as an optimal transport (OT) problem, where anchor-ground truth pairs are matched dynamically based on spatial-prioritized cost metrics.
Unlike other dynamic assignment strategies~\cite{zhang2019freeanchor, ge2021ota}, SPOTA prioritizes geometric cues by introducing normalized vertex distances and a spatial-prioritized strategy.
These designs shift focus from semantic scores to spatial alignment, which is particularly crucial in 3D scenes where geometry dominates object representation.
As a result, SPOTA establishes more stable optimization and better captures the spatial dependencies unique to 3D indoor scenes.

Then we find that dense detectors treat all positive samples equally, regardless of their relative rankings of localization accuracy or semantic reliability. 
This ranking absence prevents the model from learning to highlight good-quality predictions during training, and causes inconsistency with the rank-sensitive evaluation metric Average Precision (AP).
To resolve this challenge, we propose the Rank-aware Adaptive Self-Distillation (RAS) scheme that explicitly incorporates ranking information into the training process. 
RAS guides the classifier with localization-aware soft targets, while adaptively blended with classification loss according to confidence ranking. 
This scheme penalizes overconfident but poorly localized predictions, thereby promoting inference-aligned learning.

The main contributions of this paper are as follows:
\begin{itemize}
    \item We introduce SR3D, a highly efficient detection framework for indoor 3D object detection to mitigate the training-inference gap. Extensive experiments on ScanNet V2 and SUN RGB-D demonstrate that SR3D enhances the performance of dense detectors, while preserving real-time inference speed.
\item We present a novel Spatial-Prioritized OTA strategy, which incorporates more comprehensive geometric information and dynamically assigns labels by their qualities in a global view.
\item We design a novel Rank-aware Adaptive Self-Distillation scheme to adaptively integrate ranking awareness into the training process by self-knowledge distillation and improve compatibility with evaluation metrics.
\end{itemize}

\section{Related Work}
We here review the works about 3D object detection, dynamic label assignment and self-knowledge distillation.
\paragraph{3D Object Detection.}
We focus on indoor 3D object detection and exclude outdoor methods, which differ significantly.
Prior works are broadly categorized into sparse and dense detection paradigms.
Sparse detection methods aim to generate a limited set of high-quality proposals for object localization by the deep hough voting mechanism~\cite{qi2019deep, xie2020mlcvnet, gupta2022brnet, wang2022rbgnet, wang2022cagroup3d, zhu2024spgroup3d} or using query matching as in DETR-based methods~\cite{misra2021end, liu2021group, shen2024vdetr, wang2025state}.
In contrast, dense detection methods typically tile anchors across the spatial domain to enable dense prediction in a single shot. 
These works~\cite{gwak2020generative, rukhovich2022fcaf3d} often inherit designs from 2D frameworks~\cite{tian2019fcos}, and thus carry over limitations from 2D detection.
However, these limitations become more pronounced under the modality variations and the geometric complexity of sparse and irregular point clouds.

\paragraph{Dynamic Label Assignment.}
Label assignment, which is fundamental to 2D and 3D object detection, significantly influences the optimization of a detector, especially for dense detectors.
FreeAnchor \cite{zhang2019freeanchor} first identified the best samples based on the customized likelihood by classification scores and IoUs. 
Some other works \cite{ke2020multiple, li2020learning, zhu2020autoassign, kim2020paa} were proposed to select training samples by the joint criteria of the classification and the regression scores.
Alternatively, OTA~\cite{ge2021ota} and simOTA~\cite{ge2021yolox} handle the assignment process as an optimal transportation problem to minimize the transportation costs. 
DLLA~\cite{liu2025dlla} utilized learnable feature embedding and similarity matching to find the best assignment.
However, these methods fail to address the systematic training-inference gap stemming from the ranking awareness absence.

\paragraph{Self-Knowledge Distillation.}
Self-knowledge distillation enhances the effectiveness of training a student network by leveraging its own knowledge without an external teacher network~\cite{tommaso2018born}.
Some approaches~\cite{zhang2019your, zhu2018knowledge} introduce auxiliary networks to facilitate this process, while others~\cite{xu2019data, yun2020regularizing} adopt contrastive learning schemes to refine the internal representation learning.
Later, self-knowledge distillation has been successfully applied to various tasks such as classification~\cite{zhang2021self}, semantic segmentation~\cite{an2022efficient}, and object detection~\cite{zhang2022lgd}.
In contrast to conventional self-distillation frameworks, our approach focuses on embedding ranking awareness into the supervision process, thereby establishing a strong interaction between classification and regression branches without the need for additional modules.

\begin{figure*}[t]
    \centering
    \includegraphics[width=\linewidth]{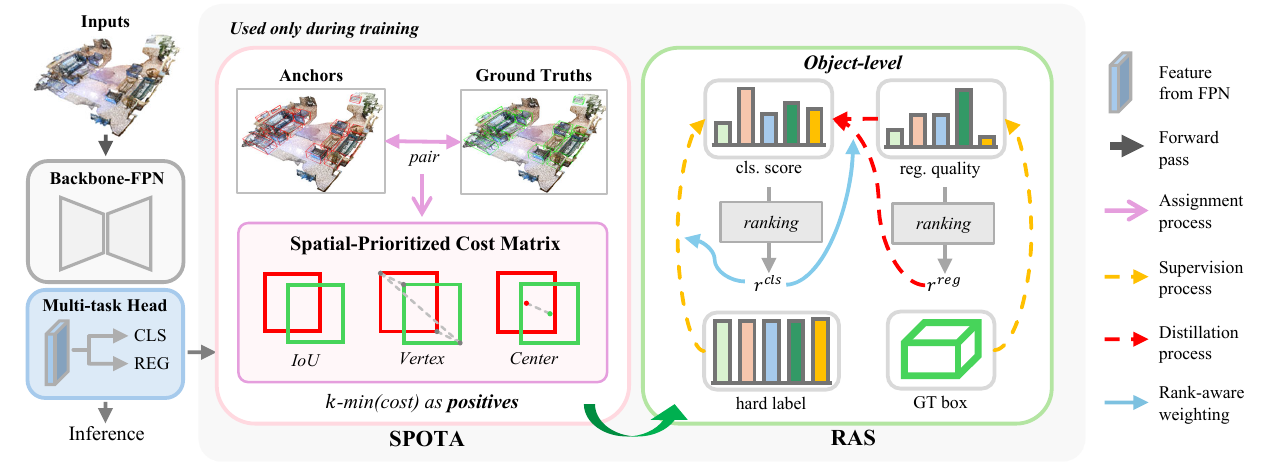}
    \caption{The overall framework of our Spatial-prioritized and Rank-aware network for indoor 3D object detection (SR3D).  
    The spatial-prioritized OTA (SPOTA) and rank-aware adaptive self-distillation (RAS) scheme are employed only during training.
    SPOTA dynamically assigns positive labels to those truly informative and high-reliability anchors by leveraging geometry hints from prediction–ground truth pairs, such as the IoU and normalized vertex distances.
    RAS introduces ranking perception into training via a self-distillation mechanism and adaptively reweights the supervision based on relative ranking signals.}
    \label{structure}
\end{figure*}

\section{Methodology}\label{methods}

In 3D object detection on point clouds, we are given a point cloud $\mathcal{S}_i$ of an indoor scene with the coordinates $\{(x, y, z)\}$ and the colors $\{(r, g, b)\}$ to produce a set of bounding boxes $\{\mathbf{b}_k\}_i$ with semantic labels $\{l_k\}_i$ to cover all objects. 

In the following sections, we first revisit the motivation behind our approach, and then detail the two core components: Spatial-Prioritized Optimal Transport Assignment (SPOTA) and Rank-aware Adaptive Self-Distillation (RAS).

\subsection{Motivation}
A key challenge in dense 3D object detection is the training-inference gap, the misalignment between how detectors are trained and how they are evaluated.
We identify two primary gaps.
(1) \textit{Missing spatial reliability}: During training, sample selection often relies on ad-hoc heuristic rules (\eg, instance scales) or prior knowledge (\eg, center prior).
These fixed methods fail to reflect the actual spatial quality of anchors, especially in cluttered indoor scenes.
(2) \textit{Missing ranking awareness}: While evaluation metrics like AP are inherently rank-sensitive, standard training pipelines apply uniform, rank-agnostic supervision across all positive samples. 
Without ranking cues, the detectors struggle to align classification confidence with true localization accuracy, leading to suboptimal results in evaluation.

To overcome these gaps, we propose SR3D, a Spatial-prioritized and Rank-aware 3D detector. SR3D tackles missing spatial reliability through Spatial-Prioritized Optimal Transport Assignment (SPOTA) and addresses missing ranking awareness via Rank-aware Adaptive Self-Distillation (RAS) during training, effectively bridging the training-inference gap in an inference cost-free manner.

The overall pipeline of SR3D is illustrated in Fig.~\ref{structure}. 
The input point clouds are processed by a sparse convolutional backbone~(Choy et al.~\citeyear{choy20194d}) with an FPN~\cite{lin2017fpn}, followed by two task-specific heads that generate dense predictions. 
During training, SR3D first assigns ground-truth labels to anchors using the proposed SPOTA, and supervises the positives with the RAS scheme. 
At inference time, Non-Maximum Suppression (NMS) is applied to remove redundant low-confidence detections.

\subsection{Preliminaries}
The Optimal Transport Assignment (OTA)~\cite{ge2021ota} is a representative and widely used label assignment method that formulates the assigning procedure as an optimal transport problem, a special form of Linear Programming (LP) in Optimization Theory.
In such a perspective, it views each ground truth ${t}_i$ as a supplier who holds $s_i$ units of positive labels, and each proposal $a_j$ as a demander who needs $d_j$ units of label.
Thus the goal of this problem is to find a transportation plan $\pi^{*} = \{ \pi_{ij}\}$, according to which all goods from suppliers can be transported to demanders at a minimal transportation cost:
\begin{equation}\small
\begin{aligned}
    \min_{\pi} \:\:\: & \sum_{i}{\sum_{j}{C_{ij}\pi_{ij}}}. \\
    \text{s.t.} \:\:\: & \sum_{i}{\pi_{ij}}=d_j, \sum_{j}{\pi_{ij}}=s_i, \\
    & \sum_i{s_i} = \sum_{j}{d_j}, \pi_{ij} \ge 0.
\end{aligned}
\end{equation}

The transportation cost from ground truth $t_i$ to proposal $a_j$ is defined as the weighted summation of their classification and regression losses in OTA:
\begin{equation}\small\label{ota-cost}
    C = \lambda \cdot \mathcal{C}_{cls} + \mathcal{C}_{reg},
\end{equation}
where $\mathcal{C}_{cls}$ and $\mathcal{C}_{reg}$ stand for cross-entropy loss and IoU loss generally, and $\lambda$ is the balancing coefficient. 

\begin{figure}[t]
    \centering
    \includegraphics[width=0.85\linewidth]{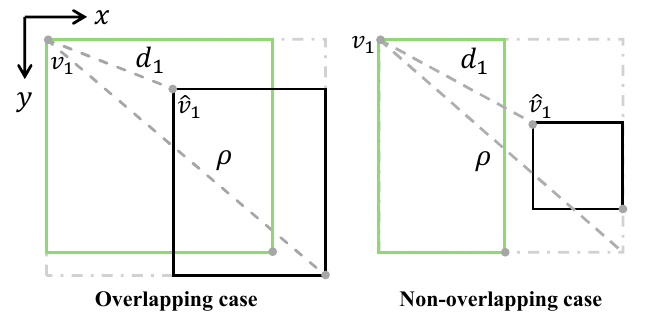}
    \caption{A simplified illustration of normalized vertex distances $\mathcal{R}_{VD}$. 
    Dashed box indicates the smallest enclosing box. We use 2D boxes for simplicity.}
    \label{vertex-distances}
\end{figure}

\subsection{Spatial-Prioritized OTA}

While OTA provides a theoretical framework introducing the awareness of prediction reliability to alleviate the missing spatial reliability, its direct application to 3D detectors suffers from several challenges:
(1) Compared to 2D object detection, 3D object detection relies more heavily on geometric information, as point cloud coordinates rather than color serve as the primary input modality.
(2) It leads to a multi-objective conflict, where predictions with either high IoU but low classification score, or vice versa, are equally likely to be selected as positives.

To overcome the limitations mentioned above, we propose a Spatial-Prioritized Optimal Transport Assignment (SPOTA). 
Compared to OTA, our proposed SPOTA integrates more geometric cues to adapt to the characteristics of 3D object detection and mitigates the multi-objective conflict by prioritizing localization quality, leading to a more coherent and stable optimization path across tasks.

First, we introduce the normalized vertex distance, a more precise and shape-sensitive measurement of spatial reliability, as shown in Fig.~\ref{vertex-distances}.
While IoU may assign similar scores to geometrically distinct predictions~\cite{re2019giou, zheng2020distance, zheng2022scaloss}, the proposed normalized vertex distance captures fine-grained differences (\eg, scale and shape) in box vertex alignment, which is defined as:
\begin{equation}\label{vd}\small
    \mathcal{R}_{VD} = \frac{d(\hat{\mathbf{v}}_1, \mathbf{v}_1) + 
    d(\hat{\mathbf{v}}_2, \mathbf{v}_2)}{2\rho(\hat{\mathbf{b}}, \mathbf{b})},
\end{equation}
where $d(\cdot)$ is the Euclidean distance, $\hat{\mathbf{v}}_1$, $\hat{\mathbf{v}}_2$ denote the vertices of the predicted box, $\mathbf{v}_1$, $\mathbf{v}_2$ are the corresponding ground truth vertices, $\rho(\hat{\mathbf{b}}, \mathbf{b})$ is the diagonal length of the smallest enclosing box covering the two boxes.
This allows SPOTA to better distinguish predictions with similar IoUs but varying spatial structures, thereby avoiding ambiguous assignments and enabling more discriminative label assignment and consistent optimization, especially in cluttered indoor environments.

Then, to mitigate the multi-objective conflict during assignment, we design a spatial-prioritized strategy for the assigning process, which is driven solely by geometric cues that better reflect the spatial nature of 3D object detection.
While classification remains essential for the recognition of 3D indoor scenes, its feature preferences and output behaviors fundamentally diverge from regression objectives, often leading to conflicting optimization signals.
Instead of merely reducing the classification cost weight by $\lambda$, we choose to completely remove it from the assignment cost.
This decision stems from the observation that, in 3D detection of point clouds, semantic cues are inherently encoded in geometric structures, \eg, object shapes, edges, and layouts~\cite{fan2024pointgcc, mei2024geo}.
Retaining an explicit classification term would thus introduce redundancy and potentially bias the assignment toward semantic patterns rather than robust geometric alignment.

Additionally, to force detectors to focus on potential positive areas and then help stabilize the training process, especially in the early stage, we impose a center prior term:
\begin{equation}\small
    \gamma_c = 1 - \exp{(- \mu d^2(\mathbf{c},\mathbf{c}^{gt}))},
\end{equation}
where $\mathbf{c}$ and $\mathbf{c}^{gt}$ are the centers of the anchor and the corresponding ground truth box, respectively.
Consequently, the final cost matrix is formulated as:
\begin{equation}\small
    C = \gamma_c \cdot (\mathcal{C}_{reg} + \mathcal{R}_{VD}),
\end{equation}
where $\mathcal{C}_{reg}$ is the regression loss, such as IoU Loss~\cite{yu2016unitbox}.
Then, following Ge et al. (2021b), we select the top $k$ predictions with the least cost as positive samples for each ground truth, while the rest are negative samples.

\subsection{Rank-aware Adaptive Self-Distillation}\label{RAD}
To resolve the issue of missing ranking awareness, we design a unified Rank-aware Adaptive Self-Distillation (RAS) paradigm, as illustrated in Fig.~\ref{structure}, which injects localization and ranking cues from the model itself during training to align supervision with inference behavior.
It consists of a self-distillation loss and an adaptive weighting strategy, both guided by the relative ranking of predictions.

We first propose a self-distillation loss to guide the classification branch with localization-aware soft targets derived from the model’s own regression branch.
Specifically, for each ground truth, we compute the localization accuracy $q$ (\ie, IoU) and the corresponding soft rank $r^{reg}$ for its positives, where higher $r^{reg}$ indicates better localization.
The rank-aware self-distillation loss is then defined as:
\begin{equation}\small
     \mathbf{RDL}(\sigma) = (1-r^{reg})^{\beta} q \log (\sigma) + q(1-q) \log(1-\sigma),
\end{equation}
where $\sigma$ is the classification confidence, and $\beta$ controls the strength of rank-based modulation.
This formulation penalizes poorly localized samples more heavily, effectively suppressing unreliable positives and encouraging the model to assign higher confidence to well-localized predictions.

Then, we need to balance the contribution of the standard classification loss and the distillation loss to stabilize training.
A fixed coefficient offers a simple way to balance the two components, but it lacks flexibility across diverse training stages and sample conditions, and fails to exploit the relative ranking crucial for optimizing AP.
To address this, we propose a rank-aware adaptive weighting mechanism that dynamically adjusts the contribution of each loss component based on the model’s own confidence ranking.
Concretely, we blend the Focal Loss~\cite{ross2017focal} and rank-aware self-distillation loss as the final classification loss:
\begin{equation}\label{cls-loss}
\small
    \mathcal{L}_{cls} = \sum_{i \in \mathcal{P}}((1- r_i^{cls}) \mathbf{FL}_i + r_i^{cls} \mathbf{RDL}_i) + \sum_{j \in \mathcal{N}} \mathbf{FL}_j,
\end{equation}
where $r_{cls}$ is the soft relative ranking of classification scores, with higher values indicating higher scores. $\mathcal{P}$ and $\mathcal{N}$ denote the sets of positive and negative anchors, respectively.

This mechanism aims to identify and rectify predictions where the model exhibits high classification confidence that is inconsistent with poor localization accuracy.
By assigning stronger distillation supervision to these potentially overconfident yet poorly localized predictions, the model is encouraged to recalibrate its internal consistency between classification confidence and localization accuracy.

\subsection{Loss Function}
The overall loss function $\mathcal{L}_{det}$ is formulated as:
\begin{equation}\small
    \mathcal{L}_{det} = \mathcal{L}_{cls} + \mathcal{L}_{reg},
\end{equation}
where $\mathcal{L}_{cls}$ is defined in Eq.~\ref{cls-loss}, and $\mathcal{L}_{reg}$ is DIoU Loss.

\begin{table*}[t]
\centering
\begin{tabular}{c | ccc | ccc }
\toprule
\multicolumn{1}{c|}{\multirow{2}{*}{Methods}}  & \multicolumn{3}{c|}{ScanNet V2} & \multicolumn{3}{c}{SUN RGB-D} \\
\multicolumn{1}{c|}{} & AP$_{25}$ $\uparrow$ & AP$_{50}$ $\uparrow$ & \multicolumn{1}{c|}{Latency $\downarrow$} & AP$_{25}$ $\uparrow$ & AP$_{50}$ $\uparrow$ & \multicolumn{1}{c}{Latency $\downarrow$} \\ 
\midrule \midrule
\multicolumn{7}{l}{\textit{Sparse detectors:}} \\
VoteNet \cite{qi2019deep} &  58.6 & 33.5 & 71ms & 57.7 & - & 41ms \\
MLCVNet \cite{xie2020mlcvnet} & 64.5 & 41.4& 183ms  & 59.8 & - & -  \\
3DETR (Misra et al. 2021) & 65.0 & 47.0 & 170ms  &  59.1 & 32.7 & - \\
BRNet~\cite{gupta2022brnet} & 66.1 & 50.9 & 135ms & 61.1 & 43.7 & - \\
GroupFree \cite{liu2021group} & 69.1 (68.6) & 52.8 (51.8) & 178ms  & 63.0 (62.6) & 45.2 (44.4) & - \\
RBGNet \cite{wang2022rbgnet} & 70.6 (69.6) & 55.2 (54.7) & 210ms  & 64.1 (63.6) & 47.2 (46.3) & -  \\
CAGroup3D \cite{wang2022cagroup3d} & 75.1 (74.5) & 61.3 (60.3) & 472ms & 66.8 (66.4) & 50.2 (49.5) & - \\
SPGroup3D \cite{zhu2024spgroup3d} & 74.3 (73.5) & 59.6 (58.3) & 131ms  & 65.4 (64.8) & 47.1 (46.4) & - \\
V-DETR \cite{shen2024vdetr} & 77.4 (76.8) & 65.0 (64.5) & 240ms & 67.5 (66.8) & 50.4 (49.7) & - \\
DEST~\cite{wang2025state} & \textbf{78.5} (78.3) & \textbf{66.6} (66.2) & 263ms & \textbf{68.4} (67.4) & \textbf{51.8} (50.9) & - \\
\midrule
\multicolumn{7}{l}{\textit{Dense detectors:}} \\
GSDN (Gwak et al. 2020) & 62.8 & 34.8 & 49ms & - & - & -\\
FCAF3D (Rukhovich et al. 2022) & 71.5 (70.7) & 57.3 (56.0) & 64ms & 64.2 (63.8)& 48.9 (48.2) & 56ms \\
\downrightarrow + DLLA~\cite{liu2025dlla} & 71.4 (71.0) & 60.0 (59.0) & 97ms & 63.8 (63.4)& 48.3 (47.4) & - \\
TR3D (Rukhovich et al. 2023) & 72.9 (72.0) & 59.3 (57.4) & \textbf{42ms} & 67.1 (66.3) & 50.4 (49.6) & \textbf{36ms} \\
\downrightarrow + DLLA~\cite{liu2025dlla} & 73.8 (72.8) & \textbf{60.2} (58.9) & - & 67.3 (67.0)& 50.6 (50.5) & - \\
\rowcolor{gray!20}
SR3D (Ours) & \textbf{74.0} (73.2) & 59.7 (58.5) & \textbf{42ms} & \textbf{68.1} (67.2) & \textbf{50.9} (50.5) & \textbf{36ms} \\
\bottomrule
\end{tabular}
\caption{Results of ours and recent indoor 3D object detection methods on the validation set of ScanNet V2 and SUN RGB-D datasets. The main comparison is based on the best results of multiple experiments between different methods, and the average value of 25 trials is given in brackets. For fair comparision, we focus on dense 3D detectors and measure the latency.}
\label{com-SOTA}
\end{table*}

\begin{figure}[t]
    \centering
    \includegraphics[width=0.9\linewidth]{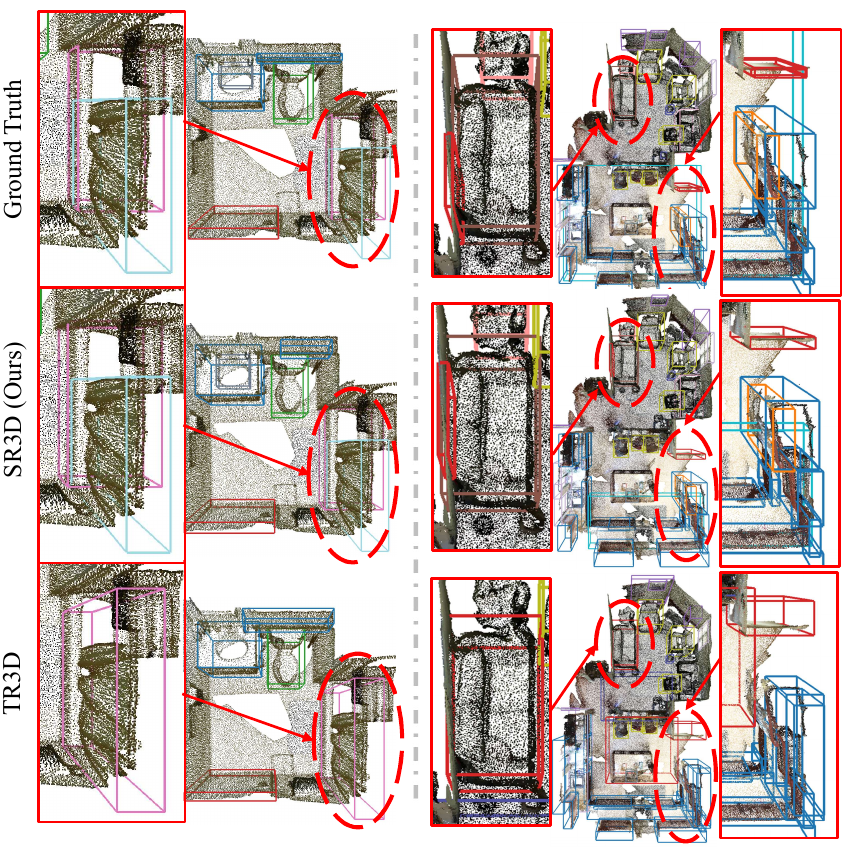}
    \caption{Qualitative results on validation set of ScanNet V2. We only visualize the most confident predictions. As compared to TR3D, our SR3D enables robust detection of more challenging objects in cluttered scenes. Different classes are indicated by bounding boxes in different colors.}
    \label{visual}
\end{figure}

\begin{figure*}[th]
    \centering
    \subfigure[]{\label{aic}\includegraphics[width=0.32\linewidth]{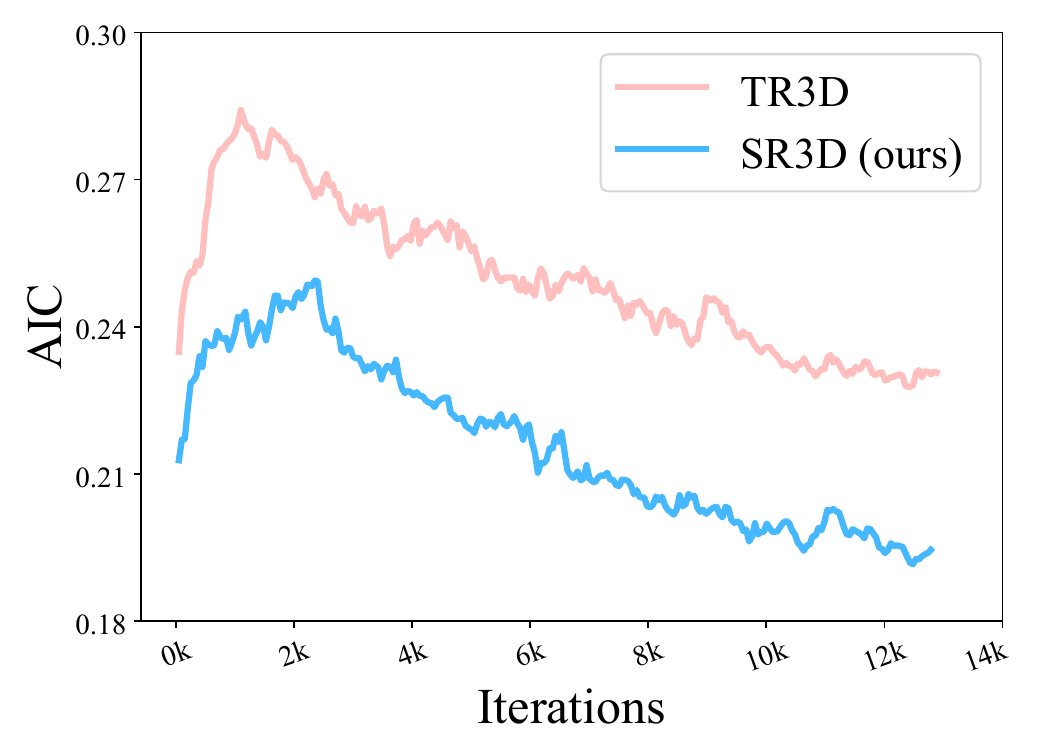}} 
    \hfill
    \subfigure[]{\label{scatter}\includegraphics[width=0.32\linewidth]{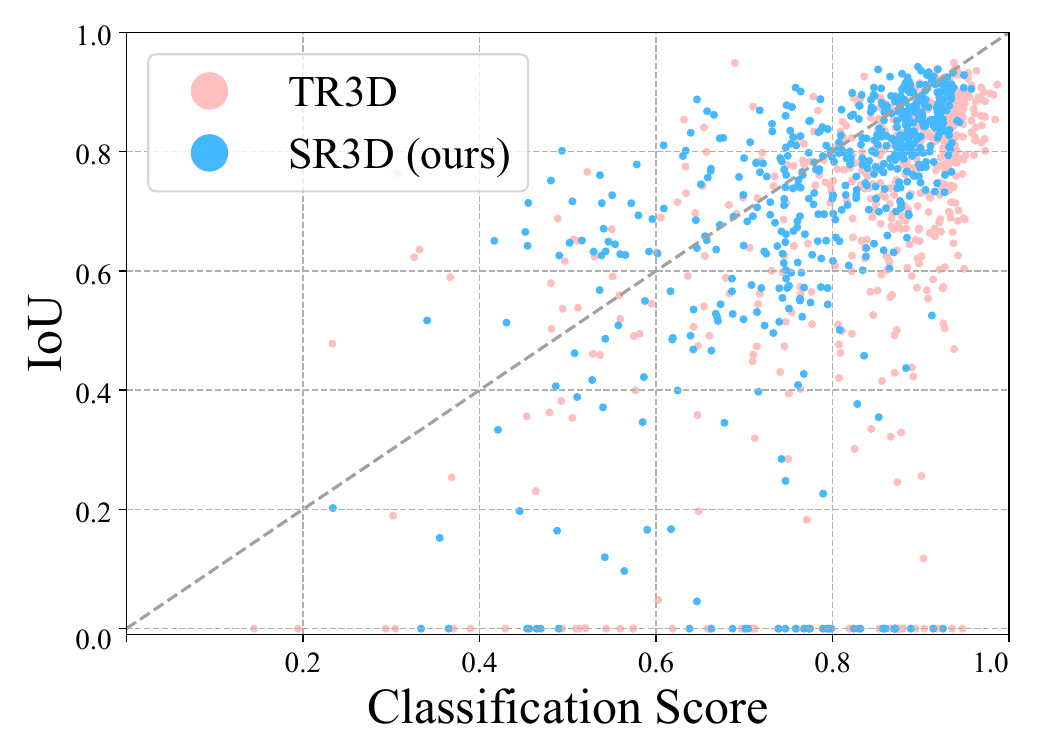}}
    \hfill
    \subfigure[]{\label{pce}\includegraphics[width=0.32\linewidth]{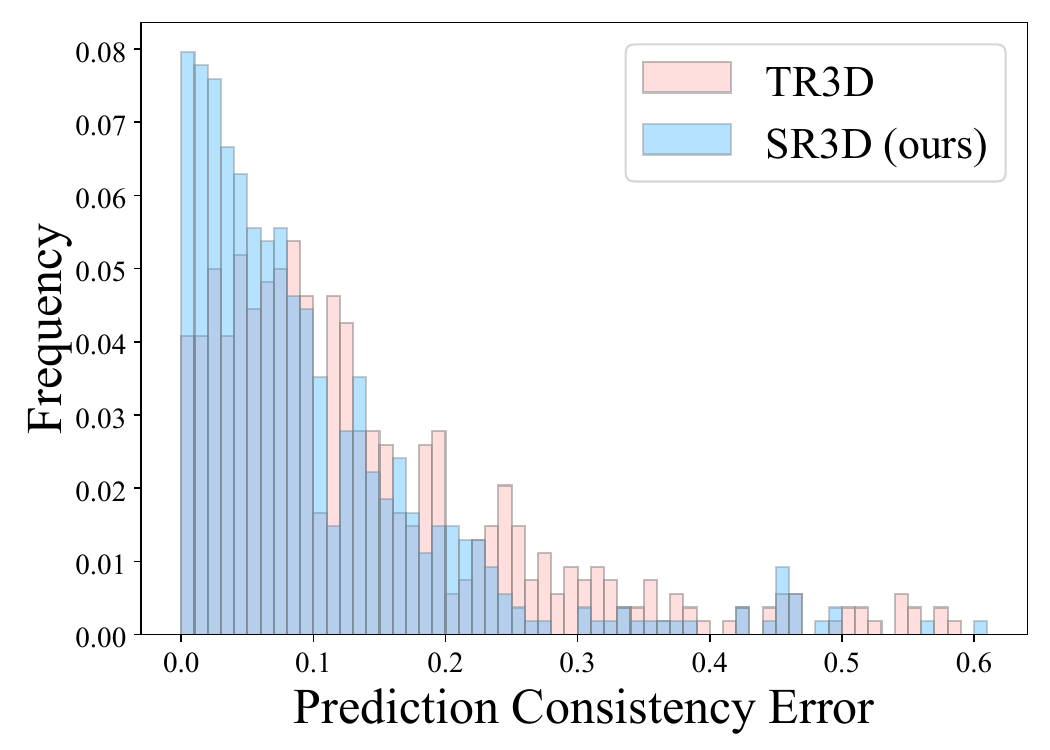}}
    \caption{The analysis of inference-aligned learning in SR3D. (a) The training Average Inconsistency Coefficient (AIC) shows lower inconsistency of SR3D. 
    (b) Confidence vs. IoU scatter plots show that the distribution of SR3D's outputs more closely with the ideal diagonal, \ie, Confidence $=$ IoU, indicating better inference consistency.
    (c) Prediction Consistency Error (PCE) reveals smaller gaps between classification scores and localization accuracy in SR3D.}
\end{figure*}

\section{Experiments}\label{experiment}

\subsection{Datasets and Evaluation Metrics}

We use two challenging 3D indoor scene datasets, ScanNet V2~\cite{dai2017scannet} and SUN RGB-D~\cite{song2015sun}, with the data splits from~\cite{qi2019deep}.

\textbf{ScanNet V2} is a richly annotated dataset that provides a comprehensive collection of 1,513 reconstructed 3D indoor scans, with per-point annotated 3D indoor scenes and bounding boxes for the 18 object categories. The dataset is divided into 1,201 training samples and 312 for validation.

\textbf{SUN RGB-D} is a widely recognized dataset designed for 3D scene understanding in indoor environments. This dataset is divided into approximately 5,285 training and 5,050 validation point clouds. 
We follow~\cite{qi2019deep} to evaluate our approach on the 10 classes of objects.

For both datasets, We follow the evaluation protocol used in 3D object detection from~\cite{qi2019deep, liu2021group}. 
Specifically, we train the model five times and test each trained model five times, yielding a total of 25 evaluation runs.
We report both the best and average values of mean Average Precision over the 25 runs, under different IoU thresholds of 0.25 (AP\textsubscript{25}) and 0.5 (AP\textsubscript{50}).

\subsection{Compared with State-of-the-arts}\label{sota}

We evaluate our SR3D with the recent state-of-the-art dense 3D detectors on ScanNet V2 \cite{dai2017scannet} and SUN RGB-D \cite{song2015sun} benchmarks.
As indicated in Tab.~\ref{com-SOTA}, SR3D outperforms the previous state-of-the-art dense detectors in all metrics, whether measured by the highest performance or the average results over multiple trials, and virtually has no effect on latency.
In terms of AP$_{25}$, our method achieves 1.1 and 1.0 improvements over the previous state-of-the-art methods on ScanNet V2 and SUN RGB-D, respectively.
While our method improves the highest AP$_{50}$ score by only 0.3 and 0.5 on ScanNet V2 and SUN RGB-D, it achieves substantial average gains of 1.1 and 0.9, suggesting that SR3D delivers more stable and reliable performance. 
SR3D and DLLA show comparable accuracy. However, DLLA suffers from higher computational overhead due to its auxiliary branch and learnable parameters. 
Our systemic inference-aligned framework is superior by achieving this performance efficiently through spatial and rank awareness, without adding any learnable components.

The visualization of 3D Object detection with predicted bounding boxes on ScanNet V2 is shown in Fig.~\ref{visual}. 
To highlight the improvements, we compare only the most confident predictions from TR3D and our SR3D, where SR3D clearly produces more accurate predictions than TR3D.

\subsection{Analysis of Inference Alignment}

To evaluate whether SR3D effectively achieves inference-aligned learning, we conduct a detailed analysis of the consistency through the following three aspects.

First, we introduce the Average Inconsistency Coefficient (AIC), defined as $AIC = {\left | \mathcal{P} \right |}^{-1} \sum_{i\in \mathcal{P}} |{p_i - q_i}|$, which measures the $L_1$ disparity between the classification score ($p_i$) and the localization quality ($q_i$).
A smaller AIC indicates better consistency.
We visualize the AIC curves during training in Fig.~\ref{aic}, and observe that SR3D consistently exhibits significantly lower AIC than TR3D, demonstrating its improved inference alignment.

Then, we examine whether this improved alignment translates into better inference behavior.
As shown in Fig.~\ref{scatter}, we visualize the top-30 high-confidence predictions of each class after NMS, plotting their classification confidence against the corresponding IoU with ground truth.

The distribution of predictions from SR3D aligns more closely with the ideal confidence–accuracy diagonal, indicating better inference consistency than TR3D.
Furthermore, we compute the absolute errors between the classification scores and IoUs of these predictions, called Prediction Consistency Error (PCE), as shown in Fig.~\ref{pce}.
Clearly, SR3D achieves lower prediction errors, further validating its ability to maintain consistent and reliable confidence outputs aligned with true localization accuracy.

These experiments collectively confirm that SR3D successfully bridges the training-inference gap in dense 3D object detection, validating the effectiveness of our inference-aligned learning designs.

\begin{table*}[t]
\begin{minipage}{0.45\textwidth}
\centering
\begin{tabular}{c c|c c c} 
\toprule
SPOTA & RAS & AP$_{25}$ & AP$_{50}$ & latency \\
\midrule
      &       & 70.8 & 55.6 & 42ms \\
\checkmark &       & 72.3 & 57.4 & 42ms \\
      & \checkmark & 72.5 & 57.7 & 42ms \\
\midrule
\rowcolor{gray!20}
\checkmark & \checkmark & \textbf{73.2} & \textbf{58.5} & 42ms \\
\bottomrule
\end{tabular}
\captionof{table}{Ablation study of SR3D on each component.}
\label{ab-study}
\end{minipage}
\hfill
\begin{minipage}{0.45\textwidth}
\centering
\begin{tabular}{l|c c}
\toprule
Setting & AP$_{25}$ & AP$_{50}$ \\
\midrule
\rowcolor{gray!20}
SPOTA (Ours) & \textbf{73.2} & \textbf{58.5} \\
w/ $\mathcal{C}_{cls}$ & 72.5 & 56.9 \\
w/o $\mathcal{R}_{VD}$ & 72.7 & 57.8 \\
\bottomrule
\end{tabular}
\captionof{table}{Effect of the designs in SPOTA.}
\label{ab-ota}
\end{minipage}
\vskip 1.5em
\begin{minipage}{0.45\textwidth}
\centering
    \begin{tabular}{c|c c}
    \toprule
    OTA-based method & AP$_{25}$ & AP$_{50}$ \\
    \midrule
       simOTA~\cite{ge2021yolox}  & 72.3 & 56.5\\
       AlignOTA~\cite{damoyolo} & 72.2 & 56.9\\
       \rowcolor{gray!20}
       SPOTA (Ours) & \textbf{73.2} & \textbf{58.5} \\
    \bottomrule
    \end{tabular}
    \caption{Superiority of SPOTA.}
    \label{Su-SPOTA}
\end{minipage}
\hfill
\begin{minipage}{0.45\textwidth}
\centering
    \begin{tabular}{c|c c}
    \toprule
        Quality-aware loss & AP$_{25}$ & AP$_{50}$ \\
        \midrule
        QFL \cite{li2020generalized} & 71.9 & 57.7\\
        VFL \cite{zhang2021varifocalnet} & 71.7 & 58.3\\
        \rowcolor{gray!20}
        RAS (Ours) & \textbf{73.2} & \textbf{58.5}\\
    \bottomrule
    \end{tabular}
    \caption{Superiority of RAS.}
    \label{su-RAS}
\end{minipage}
\end{table*}

\subsection{Ablation Study}\label{ablation}

We conduct detailed ablation studies on the validation set of ScanNet V2, reporting \textbf{\textit{the average performance}} to robustly analyze the contribution of each proposed component.

\paragraph{Effect of different components of SR3D.} 
We systematically verify the effectiveness of each component in our proposed methods on the basic fully sparse convolutional dense detector used in~(Rukhovich et al. 2023), as shown in Tab.~\ref{ab-study}. We first introduce each component individually, and the average results of all components surpass the baseline, demonstrating independent efficacy of SR3D.  
Finally, the complete model achieves 73.2 AP$_{25}$, establishing a 2.4 absolute gain over the baseline while maintaining real-time inference speed (42ms).
The detailed ablation study shows that our proposed modules consistently improve performance across various combinations, demonstrating the robustness and effectiveness of SR3D.

\paragraph{Effect of SPOTA.}

In SPOTA, we introduce several tailored designs to overcome the limitations of the standard OTA scheme. 
As shown in Tab.~\ref{ab-ota}, we conduct ablation studies to demonstrate the reasonableness of these designs.
First, reintroducing the classification loss into the cost matrix (w/ $\mathcal{C}_{cls}$) leads to a noticeable performance drop of 0.7 AP$_{25}$ and 1.6 AP$_{50}$. 
This strongly supports our spatial-prioritized strategy, demonstrating that fine-grained geometric cues are sufficient to drive effective label assignment.
Then, removing the normalized Vertex Distances regularizer (w/o $\mathcal{R}_{VD}$) causes 0.5 AP$_{25}$ and 0.7 AP$_{50}$ drop, confirming the importance of fine-grained geometric cues.
These results highlight the importance of spatial cues in cluttered 3D scenes, where spatial structure is more informative than visual appearance.

To further demonstrate the effect of SPOTA, we compare it with other recent OT-based label assignment methods, simOTA~\cite{ge2021yolox} and AlignOTA~\cite{damoyolo}.
As shown in Tab.~\ref{Su-SPOTA}, Our method achieves more reliable detection results (73.2 for AP$_{25}$ and 58.5 for AP$_{50}$) compared to these methods. 
This demonstrates that our spatial-prioritized strategy eliminates redundant influences from classification and achieves improved detection performance.

\paragraph{Effect of RAS.} 
Recent works such as Quality Focal Loss (QFL)~\cite{li2020generalized} and Varifocal Loss (VFL)~\cite{zhang2021varifocalnet} aim to embed localization quality into classification targets by directly supervising the classification branch using localization quality (\ie, IoU) as labels.
While our Rank-aware Adaptive Self-Distillation (RAS) differs in formulation from these losses, their objectives remain similar.
As shown in Tab.~\ref{su-RAS}, our RAS demonstrates superior compatibility, surpassing QFL by 1.3 for AP$_{25}$ and 0.8 for AP$_{50}$, and VFL by 1.5 for AP$_{25}$ and 0.2 for AP$_{50}$.
These results highlight a fundamental limitation of quality-aware losses in the 3D domain:
Low 3D IoU scores create optimization conflicts when localization quality is used to supervise classification, thereby harming training stability and performance.
In contrast, our RAS distills signals of localization quality and relative ranking while preserving the original objective, providing more stable and discriminative supervision, even under the low-IoU conditions common in 3D detection.

\section{Conclusion}
This paper presents SR3D, a novel and efficient framework for real-time 3D object detection of indoor scenes. 
SR3D targets the fundamental training-inference gap in dense detectors, which primarily arises from missing spatial reliability and missing ranking awareness.
We introduce two novel components to bridge this gap: the Spatial-Prioritized Optimal Transport Assignment (SPOTA) and the Rank-aware Adaptive Self-Distillation (RAS) scheme. 
The effectiveness of SR3D is validated on ScanNet V2 and SUN RGB-D, setting new benchmarks while maintaining real-time efficiency.
Extensive analysis and ablation studies further demonstrate that SR3D effectively resolves the inconsistency issue, ensuring the inference-aligned learning.

\appendix
\section{Overview}
This supplementary material provides a case study on the training-inference gap, along with additional technical details, implementation notes, and experimental results:
\begin{itemize}
    \item Sec.~\ref{case study} provides the case study to point out that the missing ranking awareness is a main bottleneck. 
    \item Sec.~\ref{td} introduces technical details about SPOTA and RAS.
    \item Sec.~\ref{id} offers more details about the implementation details of SR3D.
    \item Sec.~\ref{mr} reports the training cost, and presents more results on hyperparameters and quality visualizations, respectively.
    \item Sec.~\ref{lim} discusses the limitations of current frameworks and outlines directions for future work.
\end{itemize}

\section{Case Study}\label{case study}

To investigate the performance upper bound of the baseline model~\cite{rukhovich2023tr3d}, we replace the predicted classification score with corresponding ground-truth values for foreground points before NMS, and evaluate the performance in terms of AP on the validation set of Scannet V2~\cite{dai2017scannet}. The results are shown in Tab.~\ref{case-study}. We can see that the baseline model achieves 70.8 AP$_{25}$ and 55.6 AP$_{50}$. When using the ground-truth centerness (gt\_ctr) in inference, only about 0.1 AP$_{25}$ and 0.3 AP$_{50}$ is increased, suggesting that centerness prediction is a limiting factor but not the primary bottleneck. While replacing the classification score of the ground-truth class with the ground-truth IoU score, it achieves impressive 91.8 AP$_{25}$ and 87.7 AP$_{50}$, which are significantly higher than other cases. This indicates that the missing ranking awareness is a major bottleneck, severely limiting the model's performance. The significant gap highlights the critical need to address this inconsistency during training.

\begin{table}[b]
\centering
    \begin{tabular}{c c|c c} 
    \toprule
   gt\_ctr & gt\_iou & AP$_{25}$ & AP$_{50}$\\
      \midrule
     &  & 70.8 & 55.6 \\
    \checkmark & & 70.9 & 55.9 \\
    \rowcolor{gray!30}
     & \checkmark & \textbf{91.8} & \textbf{87.7} \\
      \bottomrule
    \end{tabular}
    \caption{Performance of baseline on ScanNet V2 with different operations. Gt\_iou means using the IoU scores with the ground-truth to replace the classification score.}\label{case-study}
\end{table}

\section{Technical Details}\label{td}
\subsection{Algorithm of SPOTA}

The detailed process of Spatial-Prioritized Optimal Transport Assignment (SPOTA) is shown in Algorithm~\ref{a-spota}. Note that we transform the vertex coordinates into polar coordinates for the prediction of the rotated bounding box, like Shen et al. (\citeyear{shen2024vdetr}).

\begin{algorithm}[t]
    \caption{Spatial-Prioritized OTA}\label{a-spota}
    \renewcommand{\algorithmicrequire}{\textbf{Input:}}
    \renewcommand{\algorithmicensure}{\textbf{Output:}}
    \begin{algorithmic}[1]
        \REQUIRE $\mathcal{G}$, $\mathcal{A}$ \\
        $\mathcal{G}$ is a set of ground-truth boxes\\
        $\mathcal{A}$ is a set of anchors
        \ENSURE $\mathcal{P}$, $\mathcal{N}$ \\
        $\mathcal{P}$ is a set of positive samples\\
        $\mathcal{N}$ is a set of negative samples
        \FOR{every ground-truth box $j \in \mathcal{G}$}
        \FOR{every anchor $i \in \mathcal{A}$}
        \STATE compute $\mathcal{R}_{VD}({i,j})$, $\mathcal{C}_{reg}(i,j)$ and $\gamma_c(i, j)$ 
        \STATE compute $C_{ij}=\gamma_c(i, j)(\mathcal{C}_{reg}(i,j)+\mathcal{R}_{VD}({i,j}))$
        \ENDFOR
        \STATE $\mathcal{P}_j$ $\leftarrow$ select top $k$ smallest $C_{ij}$ anchors
        \ENDFOR
        \STATE $\mathcal{P} = \bigcup{\mathcal{P}_j}$ 
        \STATE $\mathcal{N}=\mathcal{A} - \mathcal{P}$
    \end{algorithmic}
\end{algorithm}

\subsection{Analysis of the normalized vertex distance}

\begin{figure}[t]
    \centering
    \includegraphics[width=\linewidth]{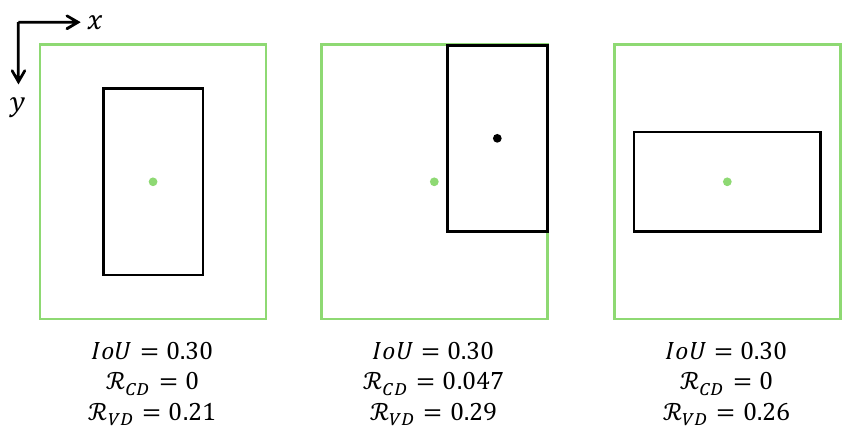}
    \caption{Compared to the normalized center distance $\mathcal{R}_{CD}$ proposed by~\cite{zheng2020distance}, our normalized vertex distance $\mathcal{R}_{VD}$ can adapt to more scenarios by incorporating shape awareness.}
    \label{vertex distance}
\end{figure}

Compared to the center-based regularizer used in DIoU Loss~\cite{zheng2020distance}, our normalized vertex distance can serve as a more comprehensive similarity measure, leading to better spatial alignment, as shown in Fig.~\ref{vertex distance}.

\subsection{Soft relative ranking}
We introduce a soft ranking algorithm~\cite{softrank} to obtain the normalized relative ranks for a sequence of scores $\{s\}$ in descending order:
\begin{equation}\label{soft-rank}
    R_i = \frac{1}{N}\sum_{j \ne i}\sigma(\frac{s_j - s_i}{\tau})
\end{equation}
where $\sigma$ is the sigmoid function, $N$ is the length of the sequence $\{s\}$, and $R \rightarrow \text{rank}(s)$ as $\tau \rightarrow 0$. 
Unlike hard ranking, which collapses the original magnitudes into discrete ordinal values and loses intra-gap variation, the soft ranking preserves the continuous pairwise distances encoded in $\{s\}$, allowing richer structural cues to be exploited.

Then, we obtain the soft rank $r$ by applying a negative exponentiation function to $R$, formulated as:
\begin{equation}
    r_i = \exp{(-R_i)}
\end{equation}
where we use $r^{reg}$ and $r^{cls}$ to present the relative ranking of IoU and classification confidence, respectively.

\section{Implementation Details}\label{id}

\subsection{Model setup details}
We use MinkResNet34 \cite{choy20194d}, a sparse 3D convolutional variant of ResNet34~\cite{he2016deep}, as the backbone feature extractor and adopt generative sparse transposed convolutions \cite{gwak2020generative} to build an FPN~\cite{lin2017fpn} for capturing multi-level features. Then, two task-specific heads process features from each layer to produce dense predictions from anchors.

\subsection{Data augmentations}

To augment the training data, we randomly sub-sample 66\% of the scene points on-the-fly. We also randomly flip the point cloud in both horizontal directions, randomly rotate the scene points by Uniform $[-5^\circ,5^\circ]$ around the upright axis, and randomly scale the points by Uniform $[0.6, 1.4]$.

\subsection{Implementation details}
The proposed detector is trained from scratch in an end-to-end manner using the MMDetection3D framework \cite{mmdet3d2020}, following the default MMDetection \cite{mmdetection} pipeline for data augmentations. 
We set the voxel size to 0.01m during initial point cloud voxelization. 
The AdamW~\cite{adamw} optimizer is used with a batch size of 16, an initial learning rate of 1e-3, which is warmed up for 300 iterations from 1e-5, and a weight decay of 1e-4. 
Training runs for 13 epochs per dataset, with the learning rate reduced by a factor of 10 at the 8th, 11th, and 12th epochs. 
At inference time, the model takes point clouds of the entire scenes and generates proposals in a forward pass. 
The proposals are post-processed through a normal 3D-NMS module for those classification confidence greater than 0.01 with an IoU threshold of 0.5, within classes.
All experiments are conducted on a single NVIDIA RTX 4090 GPU.
Following \cite{liu2021group}, each model is trained five times and tested five times per training run. 

\begin{figure}[ht]
    \centering
    \subfigure[Ablation of $k$]{\label{a}\includegraphics[width=0.48\linewidth]{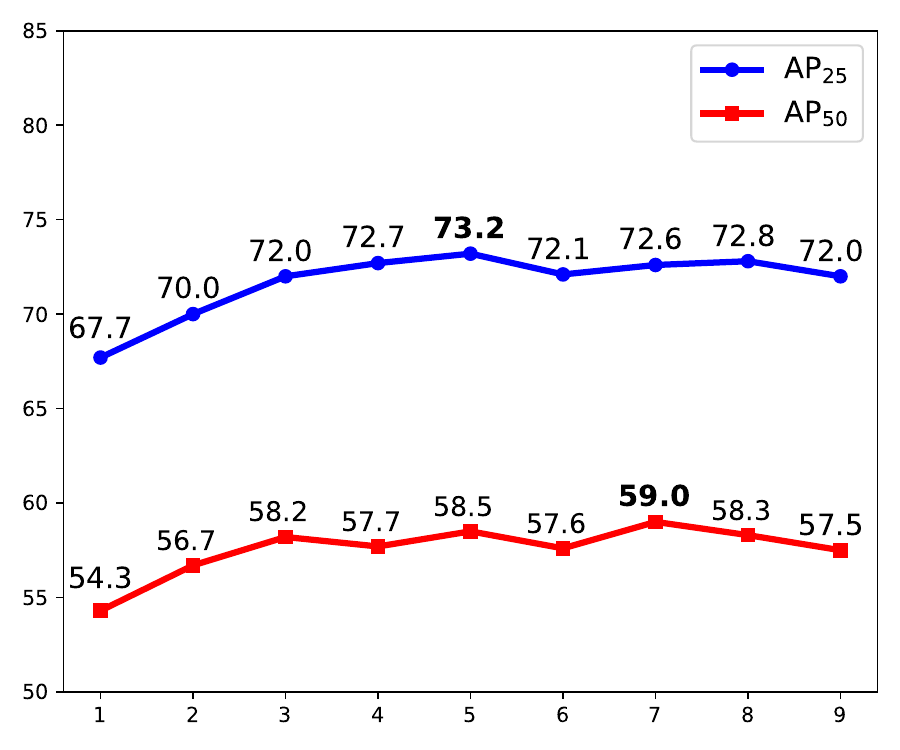}}
    \subfigure[Ablation of $\mu$]{\label{b}\includegraphics[width=0.48\linewidth]{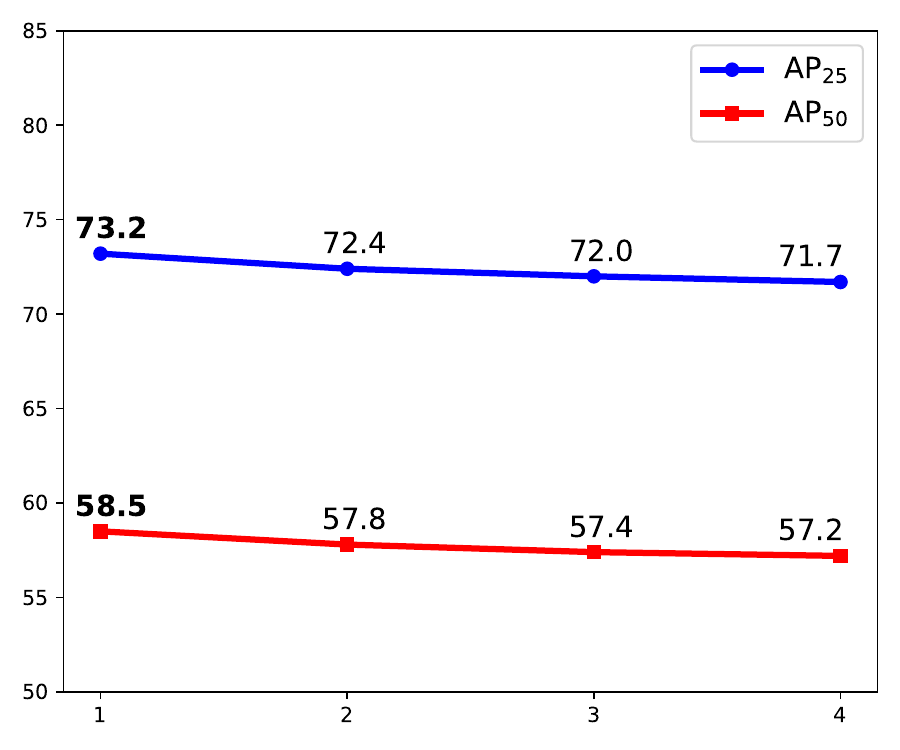}} \\
    \subfigure[Ablation of $\beta$]{\label{c}\includegraphics[width=0.48\linewidth]{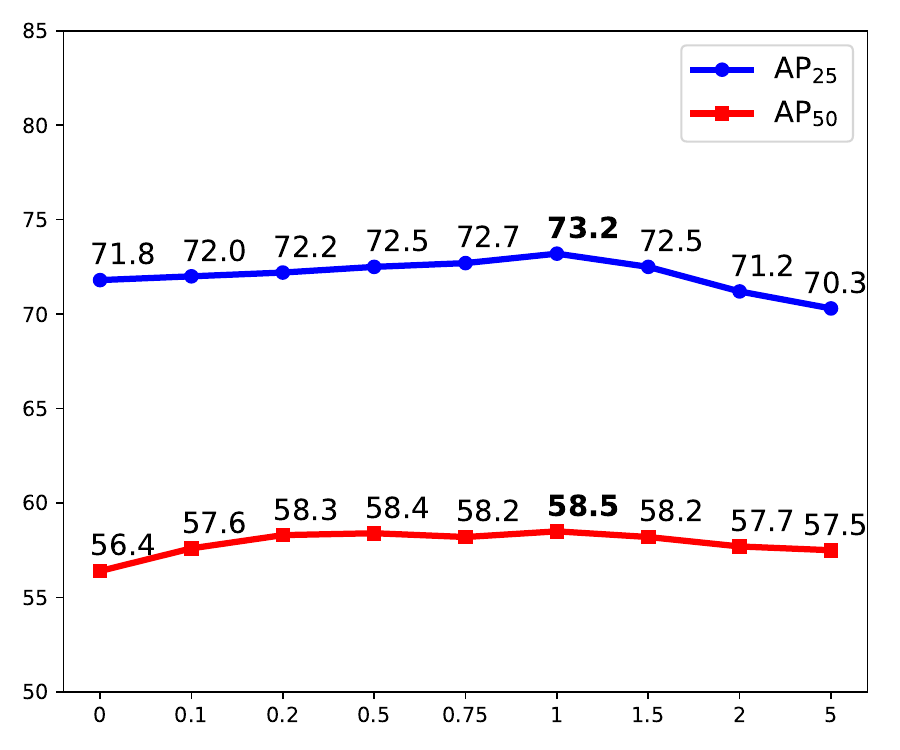}}
    \subfigure[Ablation of $\tau$]{\label{d}\includegraphics[width=0.48\linewidth]{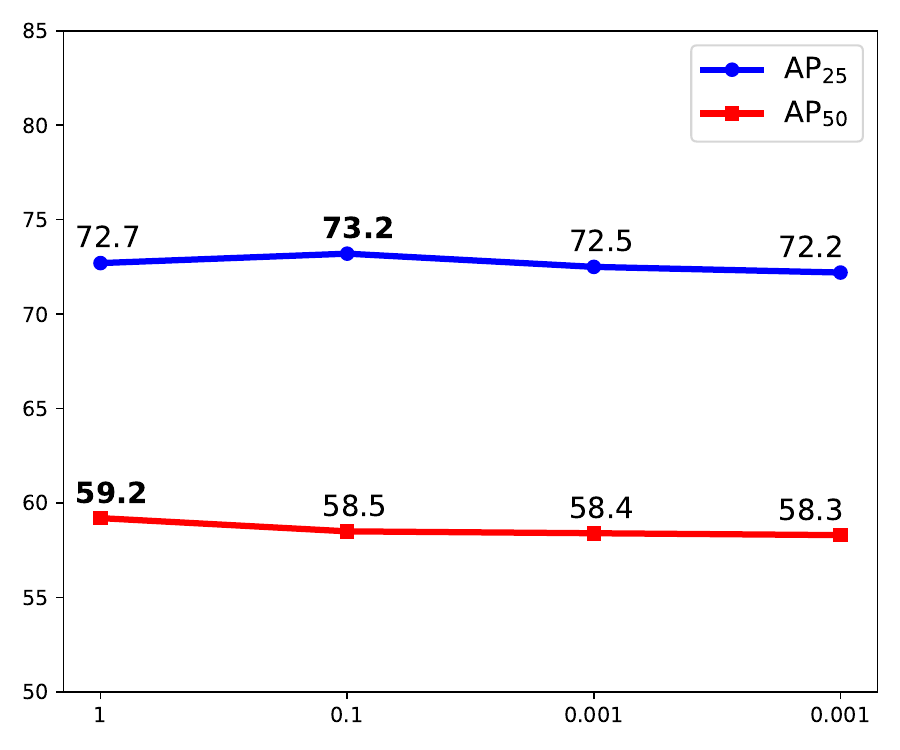}}
    \caption{Effect of hyperparameters $k$, $\mu$, $\beta$ and $\tau$ in SR3D.}\label{hyperparameter}
\end{figure}

\section{More Results}\label{mr}
In this section, we.
All experiments are conducted on a single NVIDIA RTX 4090 GPU, and we report the average performance of 25 trials by default.

\begin{table}[t]
    \centering\resizebox{\linewidth}{!}{
    \begin{tabular}{c|c c | c c}
    \toprule
    method & time & \#params & AP$_{25}$ & AP$_{50}$\\
    \midrule
        TR3D~(Rukhovich et al.~\citeyear{rukhovich2023tr3d}) & 12.3 & 14.7M & 72.0 & 57.4 \\
        SR3D (ours) & 12.6 & 14.7M & 73.2 & 58.5\\
    \bottomrule
    \end{tabular}}
    \caption{Comparison of training time (minutes) and parameters.
    The training time represents the average duration per training epoch.}
    \label{training time}
\end{table}

\subsection{Training time}

We measure the training time of different methods in Tab.~\ref{training time}. The training time denotes the average duration of each epoch. All experiments are conducted with a batch size of 16.
The results indicate that our SR3D achieves higher performance with only less than a 3\% increase in training time.

\subsection{More ablation experiments}

We provide additional ablation experiments focusing on the hyperparameters used in our model:
the candidate number $k$, the scale factor of center prior $\mu$, the rank modulate factor $\beta$ and the temperature of soft rank $\tau$.
All ablation experiments are conducted on the ScanNet V2~\cite{dai2017scannet} dataset, and we report the average performance of 25 trials by default.
The default setting chosen was $k=6$, $\mu=1$, $\beta=1$ and $\tau=0.1$.
All results are shown in Fig.~\ref{hyperparameter}

In Fig.~\ref{a}, we observe that as the number of positive candidates $k$ increases, the model achieves its highest AP$_{25}$ performance when $k=6$. When $k\ge7$, removing duplicates in the primary route becomes more difficult as the normalized soft ranking becomes smoother. Thus, the performance is unstable and achieves the highest AP$_{50}$ performance when $k=8$. Considering that $k=6$ aligns with the six principal directions of 3D Euclidean space and provides a good trade-off, we set $k=6$ by default.

As for the center factor, we observe a performance drop with $\mu$ increasing in Fig.~\ref{b}. 
This trend suggests that a larger $\mu$ places more emphasis on the center prior during label assignment, gradually degrading SPOTA into a conventional center-based scheme and weakening its spatial reliability.

For the hyperparameters used in rank-aware adaptive self-distillation (RAS), we also evaluate their effect to SR3D.
As shown in Fig~\ref{c}, the overall performance shows an upward trend as $\beta$ increases, and achieves its highest when $\beta=1$. And when $\beta$ increases further, the effect of the rank-based modulation $1-r$ is excessively weakened, leading to decreased performance.

A higher temperature $\tau$ means a smoother $r$. As illustrated in Fig.~\ref{d}, the highest AP$_{25}$ is achieved when $\tau=0.1$ and the highest AP$_{50}$ is achieved when $\tau=1$. To highlight the contribution of ranking while balancing overall performance, we set the $\tau$ to 0.01 as the default.

\subsection{More qualitative results}

\begin{figure*}[ht]
    \centering
    \includegraphics[width=\linewidth]{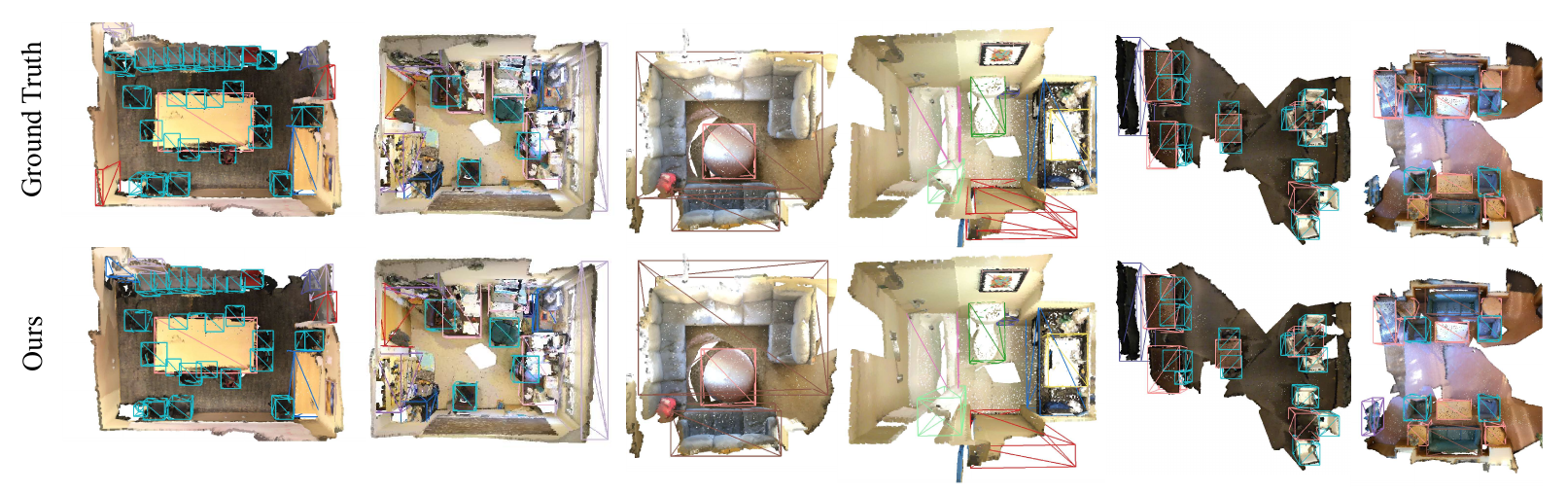}
    \caption{Qualitative results on ScanNet V2. Different classes are
indicated by bounding boxes in different colors.}
    \label{qscan}
\end{figure*}

\begin{figure*}[ht]
    \centering
    \includegraphics[width=\linewidth]{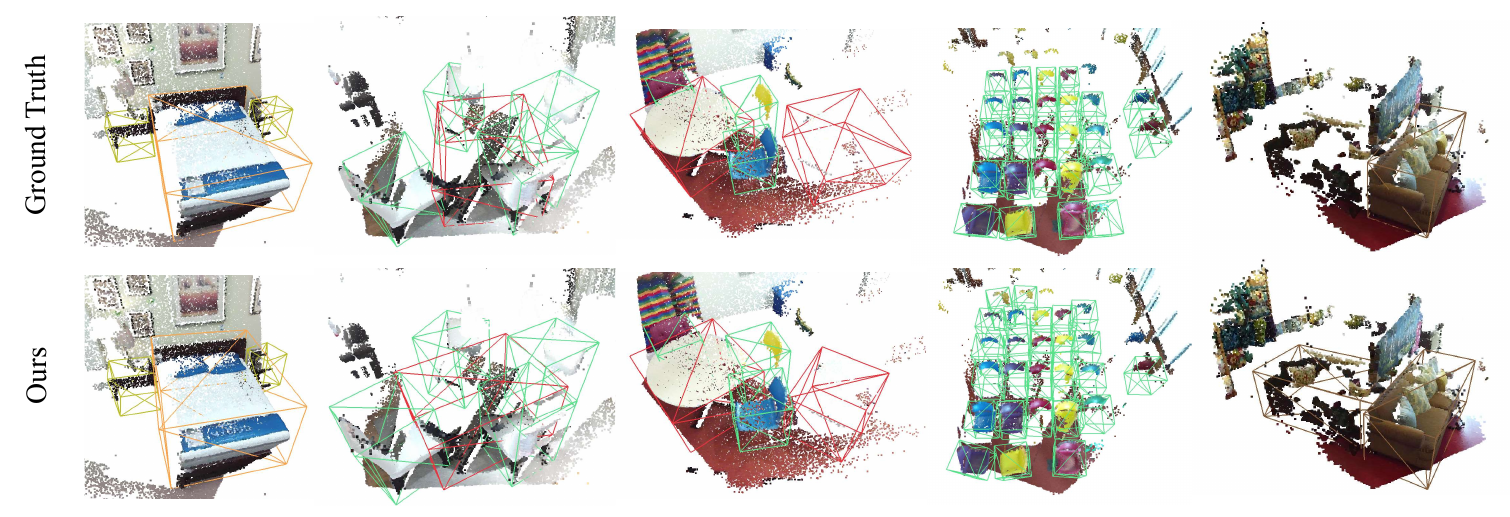}
    \caption{Qualitative results on SUN RGB-D. Different classes are
indicated by bounding boxes in different colors.}
    \label{qsun}
\end{figure*}

More qualitative results of predicted bounding boxes on the ScanNet V2 and SUN RGBD datasets are shown in Fig.~\ref{qscan} and Fig.~\ref{qsun}, respectively. In particular, from Fig.~\ref{qsun}, we can see that our proposed SR3D can even detect some unlabeled objects in SUN RGB-D.

\section{Limitations}\label{lim}

Despite the effectiveness of $\text{SR3D}$ in bridging the training-inference gap and achieving high accuracy with real-time performance on indoor 3D object detection benchmarks (ScanNet V2 and SUN RGB-D), our study presents several limitations. 
Firstly, the core design of $\text{SR3D}$ is primarily focused on enhancing model accuracy and inference consistency, but it does not specifically introduce mechanisms to improve the underlying inference speed. 
Therefore, a potential research direction is model acceleration, including further enhancing inference efficiency through techniques such as model quantization, distillation, lightweight network design, or efficient operator optimization. 
Secondly, the current validation of $\text{SR3D}$ is predominantly confined to indoor point cloud environments. 
The generalizability and efficacy of the $\text{SPOTA}$ and $\text{RAS}$ components require further investigation in large-scale indoor scenerios or more challenging outdoor autonomous driving scenarios with extreme sparsity and diverse scale distributions (e.g., $\text{LiDAR}$ data).

For future work, we plan to focus on two main areas: 
First, extending the $\text{SR3D}$ framework to handle large-scale outdoor datasets like nuScenes~\cite{Caesar2020nuscenes}, testing its robustness in complex and sparse environments.
Second, exploring model acceleration techniques to further boost $\text{SR3D}$'s computational efficiency. 
Additionally, we will investigate the potential of multimodal fusion (e.g., incorporating $\text{RGB}$ images and texts) and how the concept of inference-aligned learning can further elevate the performance of multimodal 3D object detectors.

\section*{Acknowledgments}
This research is supported by the National Key Research and Development Program of China 2024YFC3811000, the NSFC-projects 42471447, and the Fundamental Research Funds for the Central Universities of China 2042022dx0001.

\bibliography{aaai2026}

\end{document}